\pdfoutput=1
\documentclass[runningheads]{llncs}
\usepackage[T1]{fontenc}
\usepackage{graphicx}
\usepackage{amssymb}
\usepackage{amsmath}

\begin{document}
\title{AI-Driven Surrogate Models for Predicting Electrode-Scale Discharge Behavior in Lithium-Ion Batteries}
\titlerunning{AI-Driven Surrogate Models for LIB Discharge Prediction}
\author{Mengda Xing\inst{1}\orcidID{0000-0001-9838-4687} \and
    Jean-Marie Lagniez\inst{1}\orcidID{0000-0002-6557-4115} \and
    Alejandro A. Franco\inst{2}\orcidID{0000-0001-7362-7849}}
\authorrunning{M. Xing et al.}

\institute{CRIL, UMR 8188, Universit\'{e} d'Artois,\\
Rue Jean Souvraz SP 18, F-62307 Lens Cedex, France\\
\email{\{xing,lagniez\}@cril.fr}
\and
LRCS, UMR 7314, Universit\'{e} de Picardie Jules Verne, \\
15 rue Baudelocque, 80039 Amiens Cedex, France\\
\email{alejandro.franco@u-picardie.fr}}

\maketitle              
\begin{abstract}
    Physics-based simulations are essential for understanding the electrode-scale discharge behavior of lithium-ion batteries (LIBs) but suffer from prohibitive computational costs. To address this, we introduce a novel deep learning surrogate pipeline based on the Swin3D Transformer to predict spatiotemporal discharge dynamics directly from volumetric data. Our approach integrates two key innovations: Gaussian Positional Encoding (GPE), which enhances spatial feature representation by adapting to the complex geometry of electrode microstructures, and a specialized Temporal Encoding module to capture non-linear time-series evolution. Experimental validation on an Electrochemical Simulation (ES) dataset demonstrates that our pipeline significantly outperforms state-of-the-art point cloud baselines in prediction accuracy. Furthermore, the proposed method reduces the computational overhead by orders of magnitude, providing a scalable and efficient framework for high-throughput battery design and optimization.
\end{abstract}
\section{Introduction}

Physics-based simulations using Finite Element Methods (FEM) are essential for analyzing Lithium-Ion Battery (LIB) microstructures but suffer from prohibitive computational costs, often requiring hours per discharge cycle \cite{duquesnoy2023machine}. While Machine Learning (ML) surrogates offer acceleration, integrating them into legacy simulation frameworks (e.g., COMSOL, custom Fortran/C++ solvers) presents significant software engineering challenges. These monolithic systems lack the modularity to ingest modern AI components, making manual integration error-prone and hindering the adoption of "AI for Science" paradigms.

\begin{figure}[t]
    \centering
    \includegraphics[width=0.9\textwidth]{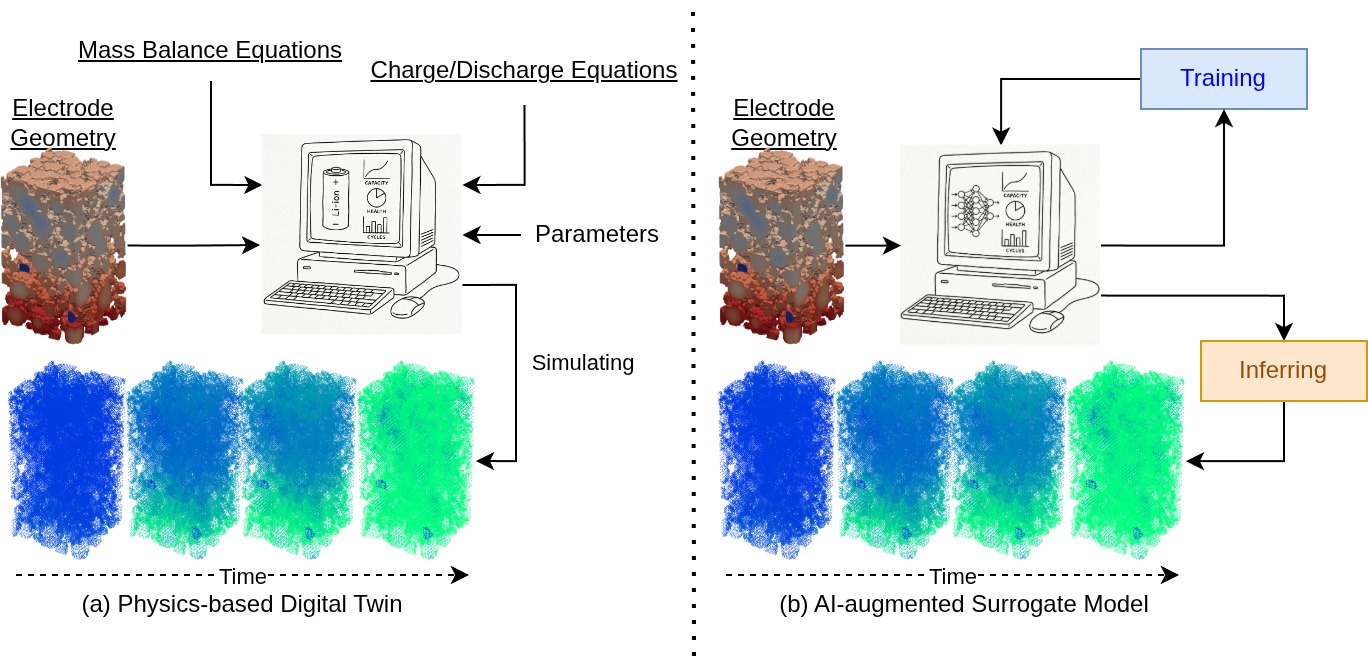}
    \caption{Transition from (a) traditional numerical simulation to (b) the proposed AI-augmented surrogate modeling framework.}
    \label{fig:intro_framework}
\end{figure}

To address this, we propose an automated framework that refactors rigid physics-based pipelines into hybrid, AI-augmented architectures. As illustrated in Fig.~\ref{fig:intro_framework}, our approach systematically replaces computationally intensive numerical solvers with deep learning surrogates. By treating AI adoption as a software refactoring process, we enable domain experts to evolve legacy systems with minimal manual intervention.

Technically, we employ Swin3D Transformers enhanced with Gaussian Positional Encoding to learn spatiotemporal dynamics directly from 3D unstructured mesh data. This pipeline automates the end-to-end workflow from data extraction to model deployment ensuring physical consistency while achieving orders-of-magnitude acceleration.

The main contributions of this work are:
\begin{itemize}
    \item \textbf{Automated Refactoring Framework:} We introduce a pipeline that modernizes legacy battery simulation software by systematically integrating AI surrogates, improving both maintainability and efficiency.
    \item \textbf{Swin3D Surrogate Model:} We propose a specialized deep learning architecture that directly processes volumetric mesh data, serving as a high-fidelity replacement for traditional solvers.
    \item \textbf{Efficiency Breakthrough:} Validated on a large-scale Electrochemical Simulation (ES) dataset, our approach reduces simulation time from hours to milliseconds without compromising accuracy.
\end{itemize}
\section{Methodology}

\begin{figure}[t]
    \centering
    \includegraphics[width=\textwidth]{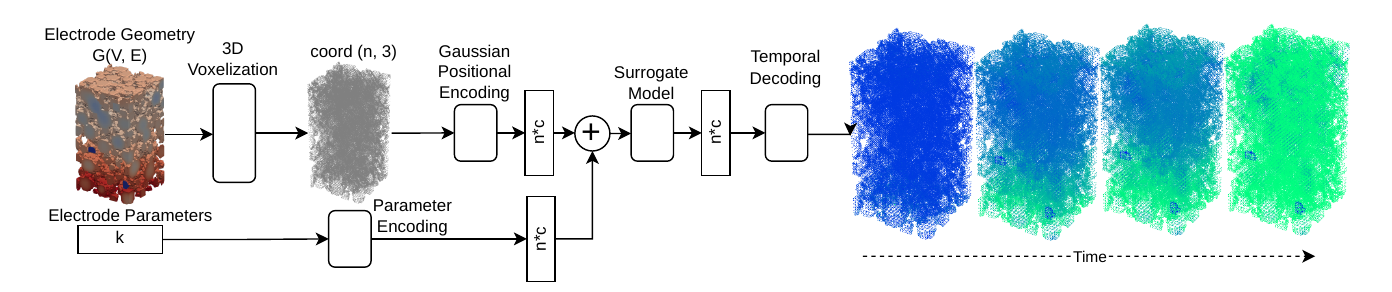}
    \caption{Overview of the proposed point cloud forecasting pipeline. The system takes electrode geometry (red) as input, processes it through voxelization and a Swin3D-based surrogate model, and predicts the spatiotemporal discharge behavior (blue-to-green).}
    \label{fig:method}
\end{figure}

As illustrated in Fig.~\ref{fig:method}, our pipeline predicts the spatiotemporal discharge behavior of LIB cathodes. The framework consists of three stages: (1) converting unstructured mesh data into sparse voxel grids; (2) extracting geometric features using Gaussian Positional Encoding (GPE) and a Swin3D backbone; and (3) decoding temporal dynamics to generate time-series predictions.

\subsection{Data Representation}
The input data is derived from physics-based simulations of electrode microstructures \cite{LIU2023156}. We treat the finite element mesh nodes as a point cloud $P = \{x_i\}_{i=1}^N \in \mathbb{R}^{N \times 3}$.
To handle the high density of simulation points (approx. $10^5$ points per sample) efficiently, we employ 3D Voxelization. This process quantizes the continuous coordinates into a sparse voxel grid, reducing computational complexity while preserving the macroscopic geometric topology required for the Swin3D backbone.

\subsection{Network Architecture}

\subsubsection{Gaussian Positional Encoding (GPE).}
To compensate for the quantization loss in voxelization and capture sub-voxel geometry, we introduce GPE. Unlike static sine-cosine encodings, GPE utilizes learnable Gaussian kernels to map coordinates into high-dimensional feature spaces, effectively modeling complex non-Euclidean structures \cite{chandra2021survey}.

\begin{figure}[t]
    \centering
    \includegraphics[width=0.6\linewidth]{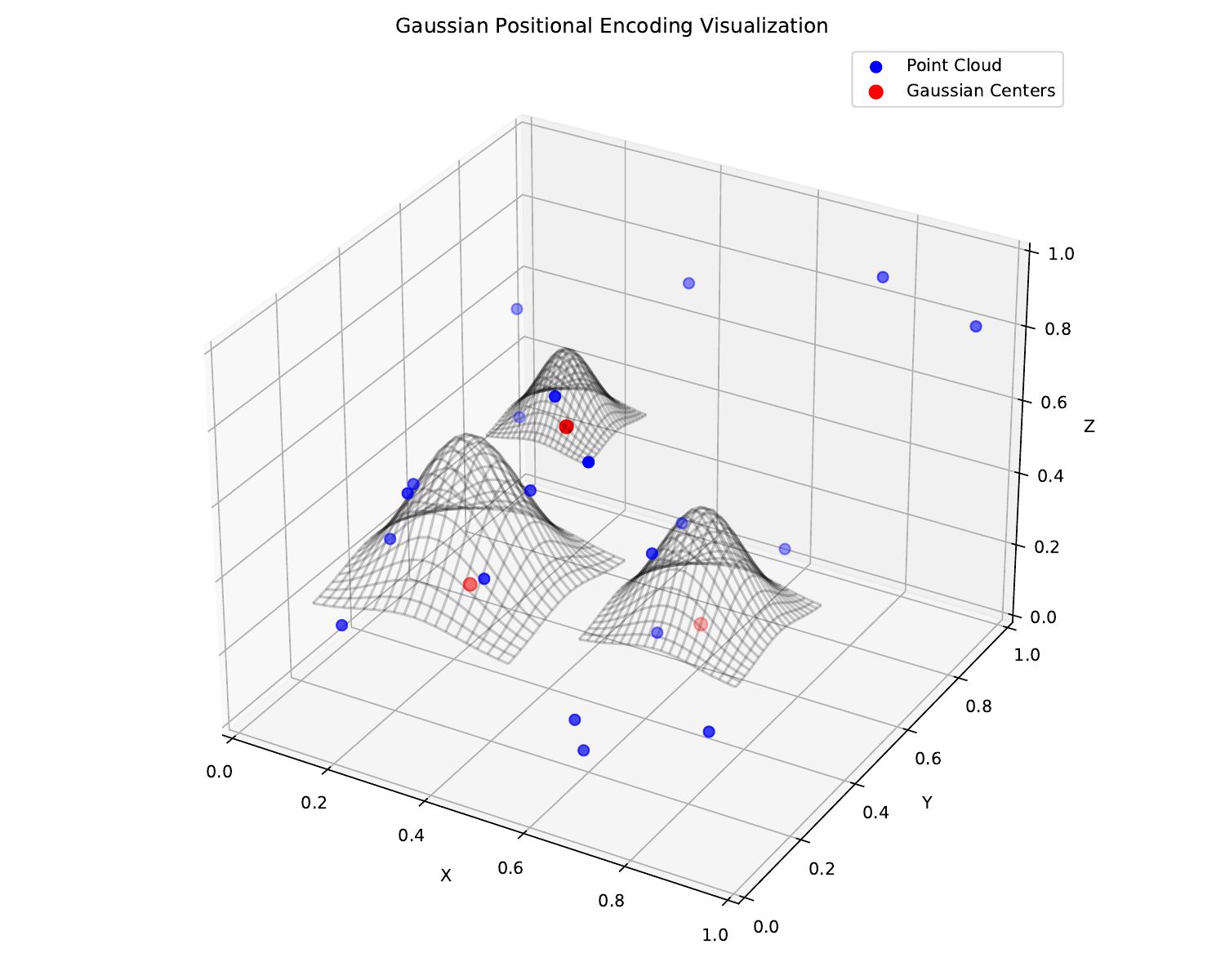}
    \caption{Gaussian Positional Encoding. Blue points: input coordinates; Red points: learnable Gaussian kernel centers.}
    \label{fig:gaussian-positional-encoding}
\end{figure}

As defined in Eq.~\ref{eq:gpe}, for a point $\mathbf{x} \in \mathbb{R}^3$ and a kernel with center $\boldsymbol{\mu}$ and bandwidth $\sigma$:
\begin{equation}\label{eq:gpe}
    \mathcal{G}(\mathbf{x}, \boldsymbol{\mu}, \sigma)=\exp \left(-\frac{\|\mathbf{x}-\boldsymbol{\mu}\|^2}{2 \sigma^2}\right)
\end{equation}
Both $\boldsymbol{\mu}$ and $\sigma$ are learnable parameters. We stack outputs from $K$ kernels to form the encoding vector $\operatorname{GPE}(\mathbf{x}) \in \mathbb{R}^K$.

\subsubsection{Surrogate Model \& Temporal Decoding.}
\begin{figure}[t]
    \centering
    \includegraphics[width=\textwidth]{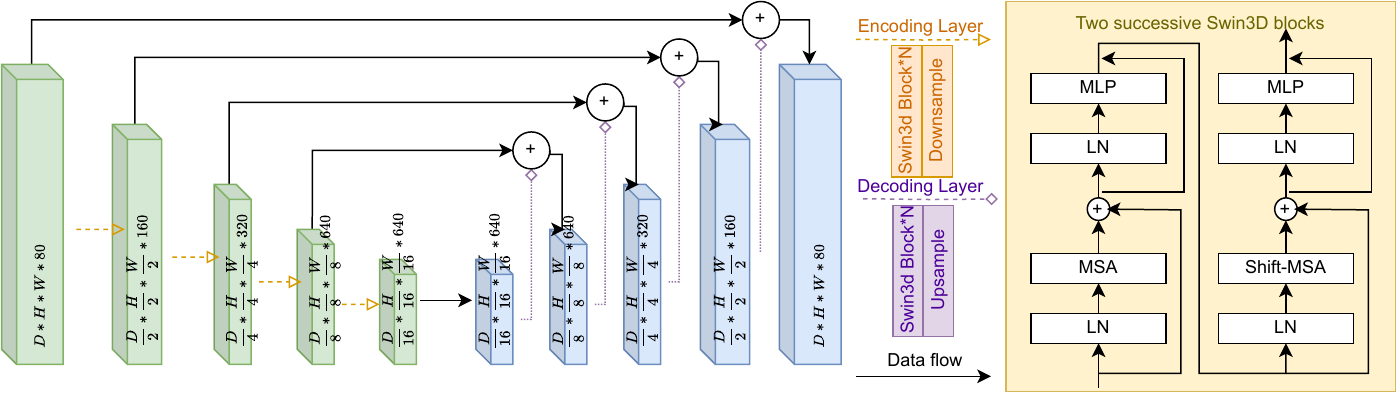}
    \caption{Left: The Swin3D backbone architecture. Right: Details of two successive Swin3D Blocks with shifted windows.}
    \label{fig:swin3d}
\end{figure}

We utilize the Swin3D Transformer (Fig.~\ref{fig:swin3d}) as the backbone. By employing shifted-window Multi-head Self-Attention (MSA), Swin3D efficiently captures both local fine-grained details and global semantic contexts within the voxel grid.
To predict the time-series evolution (24 time steps), we append a Temporal Encoding module, implemented as a Multi-Layer Perceptron (MLP), to the backbone's output.

\subsubsection{Optimization.}
The network is trained in an end-to-end manner using the Mean Squared Error (MSE) loss: $\mathcal{L} = \frac{1}{n} \sum (Y_{gt} - \hat{Y}_{pred})^2$, aiming to minimize the discrepancy between the predicted electrochemical fields and the physics-based ground truth.

\section{Related Work}
\label{sec:related_work}

We review existing literature on 3D deep learning, categorizing methods into two primary streams: point-based architectures and transformer-based architectures.

\subsubsection{Point-based Methods.}
PointNet \cite{qi2017pointnet} pioneered the direct processing of unordered point clouds using deep learning. It employs a shared Multi-Layer Perceptron (MLP) to process each point independently, followed by a symmetric max-pooling function to aggregate global features. While efficient, PointNet struggles to capture local geometric structures due to its lack of local context awareness.
To address this, PointNet++ \cite{qi2017pointnet++} introduced a hierarchical learning mechanism that recursively applies PointNet to local neighborhoods, enabling the extraction of fine-grained geometric details at multiple scales.
Further improving local feature learning, DGCNN \cite{wang2019dynamic} proposed the EdgeConv operation, which constructs dynamic k-nearest neighbor ($k$-NN) graphs in the feature space. This allows the network to explicitly model the relationship between a point and its neighbors, learning geometry-aware embeddings that capture the topological structure of the data.

\subsubsection{Transformer-based Methods.}
Inspired by the success of Transformers in natural language processing and 2D vision, recent works have adapted self-attention mechanisms to 3D data.
Point Transformer \cite{zhao2021point} introduces a vector self-attention block specifically designed for point clouds. By leveraging positional encoding and attention within local neighborhoods, it effectively captures both geometric structures and global dependencies, outperforming traditional MLP-based methods on semantic segmentation tasks.
Building on this, Point Transformer V2 (PTv2) \cite{wu2022point} incorporates grouping vector attention and partition-based pooling to further enhance computational efficiency and feature integration.
Moving beyond supervised learning, Point-BERT \cite{yu2022point} introduces a masked point modeling paradigm, adapting the BERT pre-training framework to learning generic 3D representations via masked auto-encoding.

Most relevant to our work is the Swin3D Transformer \cite{yang2023swin3d}. Building upon the hierarchical design of the Swin Transformer \cite{liu2021swin}, Swin3D extends the shifted window attention mechanism to the 3D domain. Unlike point-based methods that operate on unstructured coordinates, Swin3D processes voxelized volumetric data, making it particularly effective for tasks requiring structured spatial understanding, such as medical image analysis and large-scale scene understanding \cite{xue2022urban,siddiqui2023panoptic}. This ability to handle structured 3D grids makes it an ideal backbone for predicting dense physical fields in battery simulations.
\section{Experiments}

\subsection{Experimental Setup}
The surrogate models were implemented using PyTorch 2.3.0 on a workstation running Ubuntu 22.04. The hardware configuration includes an NVIDIA RTX A4500 GPU (20 GB VRAM) and 64 GB of RAM. All training and inference processes were GPU-accelerated. The physics-based ground truth data was generated using COMSOL Multiphysics, a standard finite element analysis solver for electrochemical modeling.

\subsection{Experimental Settings}

\subsubsection{Baselines.}
To evaluate the effectiveness of our proposed method, we compare \texttt{Swin3D} against representative state-of-the-art point cloud learning architectures.
For point-based methods, we select PointNet \cite{qi2017pointnet}, which processes points independently, and DGCNN \cite{wang2019dynamic}, which utilizes dynamic graph convolutions.
For transformer-based methods, we compare against Point Transformer (PT) \cite{zhao2021point} and Point Transformer V2 (PTv2) \cite{wu2022point}, which apply self-attention mechanisms to point sets.
All baselines were retrained on the ES dataset using identical protocols to ensure a fair comparison.

\subsubsection{Model Variants.}
We analyze the impact of positional encoding and model capacity through various configurations:
\begin{itemize}
    \item \textbf{Positional Encodings:} We compare our proposed Gaussian Positional Encoding (GPE) against Sine-Cosine (SIN) encoding and raw coordinates (XYZ). Unlike fixed SIN functions, GPE employs learnable Gaussian kernels that dynamically adapt to the spatial distribution of the electrode microstructures, thereby capturing intricate geometric details more effectively.
    \item \textbf{Model Capacity:} We evaluate four scales of the Swin3D architecture: Tiny (T), Small (S), Medium (M), and Large (L). The channel configurations for the five stages are defined as follows:
          \begin{itemize}
              \item \texttt{Swin3D-T}: $C = \{32, 64, 128, 256, 256\}$
              \item \texttt{Swin3D-S}: $C = \{48, 96, 192, 384, 384\}$
              \item \texttt{Swin3D-M}: $C = \{64, 128, 256, 512, 512\}$
              \item \texttt{Swin3D-L}: $C = \{80, 160, 320, 640, 640\}$
          \end{itemize}
\end{itemize}
Smaller variants (T, S) prioritize computational efficiency, while larger variants (M, L) offer higher theoretical capacity.

\subsubsection{Dataset.}
The summary of the dataset is shown in Table \ref{tab:dataset_summary}.
We validate our pipeline on the Electrochemical Simulation (ES) dataset, which models the discharge process of an NMC111 cathode. The dataset contains high-fidelity 3D fields of lithium concentration and electric potential within the active material and carbon binder domains.

\begin{table}[t]
    \centering
    \caption{Summary of the Electrochemical Simulation (ES) Dataset. The dataset covers varying electrode designs and operational conditions, providing dense 3D concentration fields over time.}
    \label{tab:dataset_summary}
    \begin{tabular}{ll}
        \hline
        \textbf{Attribute}                 & \textbf{Specification}                            \\
        \hline
        \multicolumn{2}{l}{\textit{Geometry \& Scale}}                                         \\
        Dimensions ($W \times D \times H$) & $34.5 \times 34.5 \times 123$ (unitless)          \\
        Number of Points                   & $9.4 \times 10^5 \sim 2.4 \times 10^6$ per sample \\
        Number of Cells (Tetrahedra)       & $5.5 \times 10^5 \sim 7.4 \times 10^5$ per sample \\
        \hline
        \multicolumn{2}{l}{\textit{Simulation Parameters}}                                     \\
        Active Material (AM)               & 87\%, 93\%, 95\% (wt)                             \\
        Calendering Degree (Cal)           & 0\%, 10\%, 20\%                                   \\
        Discharge Rate (C-rate)            & 0.5C, 1C, 2C                                      \\
        \hline
        \multicolumn{2}{l}{\textit{Data Structure}}                                            \\
        Time Steps                         & 24 steps ($t=0$ to $t=23$)                        \\
        Target Variable                    & Li-Ion Concentration ($C$)                        \\
        Total Samples                      & 18 (16 Training, 2 Testing)                       \\
        \hline
    \end{tabular}
\end{table}

\subsection{Evaluation Metrics}

We employ both pointwise error maps for visual assessment and aggregated statistical metrics for numerical comparison.

\textbf{Visual Metrics.} To visualize the spatial distribution of errors, we define the Pointwise Error (PE) as the difference between the prediction $\hat{Y}$ and the ground truth $Y$. The Mean Absolute Error (MAE) aggregates this over all $n$ points:
\begin{equation}
    \text{PE} = \hat{Y} - Y, \quad \quad \text{MAE} = \frac{1}{n} \sum_{i=1}^{n}{|\hat{Y}_i - Y_i|}
\end{equation}
Note that in the results section, we visualize PE to identify regions of under- or over-estimation.

\textbf{Numerical Metrics.} We use three standard metrics to quantify overall performance: Root Mean Squared Error (RMSE) for error magnitude, Symmetric Mean Absolute Percentage Error (SMAPE) for relative accuracy, and the Coefficient of Determination ($R^2$) for explanatory power:

\begin{equation}
    \text{RMSE} = \sqrt{\frac{1}{n} \sum_{i=1}^{n} \left( Y_i - \hat{Y}_i \right)^2}
\end{equation}

\begin{equation}
    \text{SMAPE} = \frac{100\%}{n} \sum_{i=1}^{n} \frac{|\hat{Y}_i - Y_i|}{(|\hat{Y}_i| + |Y_i|)/2}
\end{equation}

\begin{equation}
    R^2 = 1 - \frac{\sum_{i=1}^{n} (Y_i - \hat{Y}_i)^2}{\sum_{i=1}^{n} (Y_i - \bar{Y})^2}
\end{equation}
where $\bar{Y}$ represents the mean of the observed data. Higher $R^2$ and lower RMSE/SMAPE indicate better performance.
\section{Results}

In this section, we evaluate the performance of our proposed method on the electrochemical simulation dataset. We present a comprehensive analysis including visual quality assessment, quantitative numerical comparison, computational efficiency analysis, and ablation studies on the key components: Gaussian Positional Encoding (GPE) and Temporal Encoding.

\subsection{Visual Comparison}

\begin{figure}[t]
    \includegraphics[width=\textwidth]{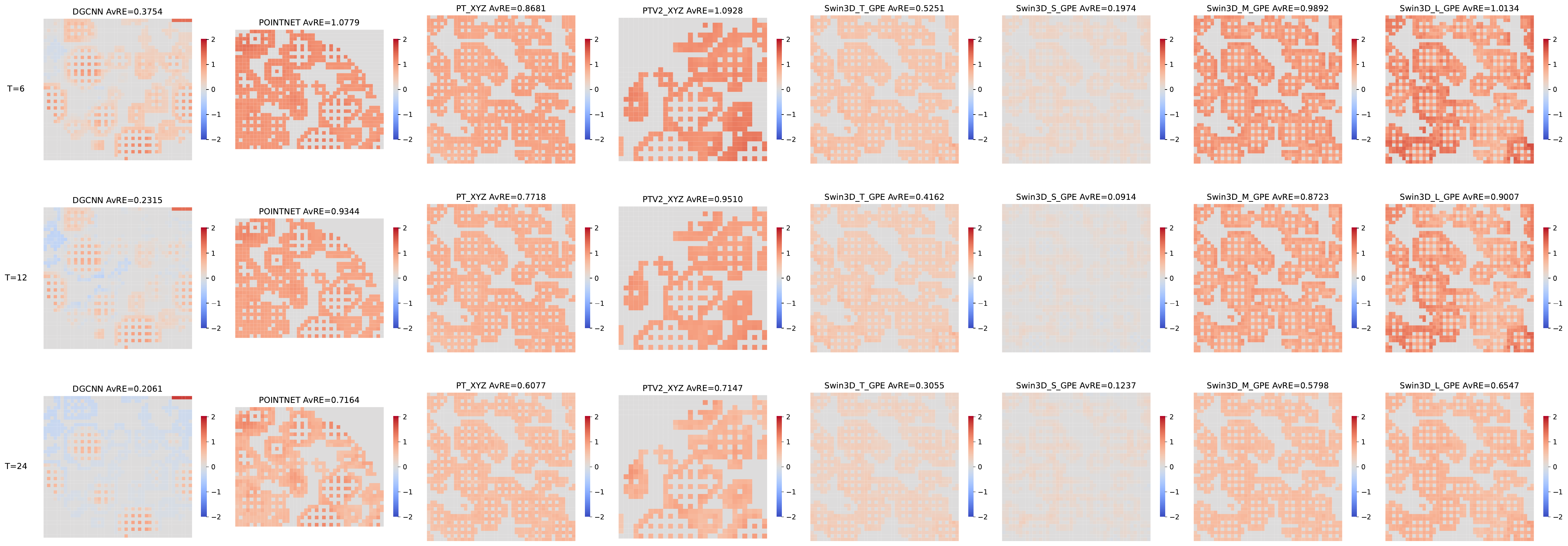}
    \caption{Visual comparison of the Relative Error (RE) in lithium-ion concentration predictions. The results are visualized as 2D slices extracted from the 3D volumetric predictions. Each column corresponds to a different model, and each row represents a specific time step ($t = 6, 12, 24$). Ideally, the error map should be gray (indicating zero error).}
    \label{fig:loss_baseline}
\end{figure}

We first assess the visual fidelity of the lithium-ion concentration predictions. Fig.~\ref{fig:loss_baseline} illustrates the spatial distribution of the Relative Error (RE) across different time steps ($t = 6, 12, 24$).

\begin{figure}[t]
    \includegraphics[width=\textwidth, height=0.45\textheight, keepaspectratio]{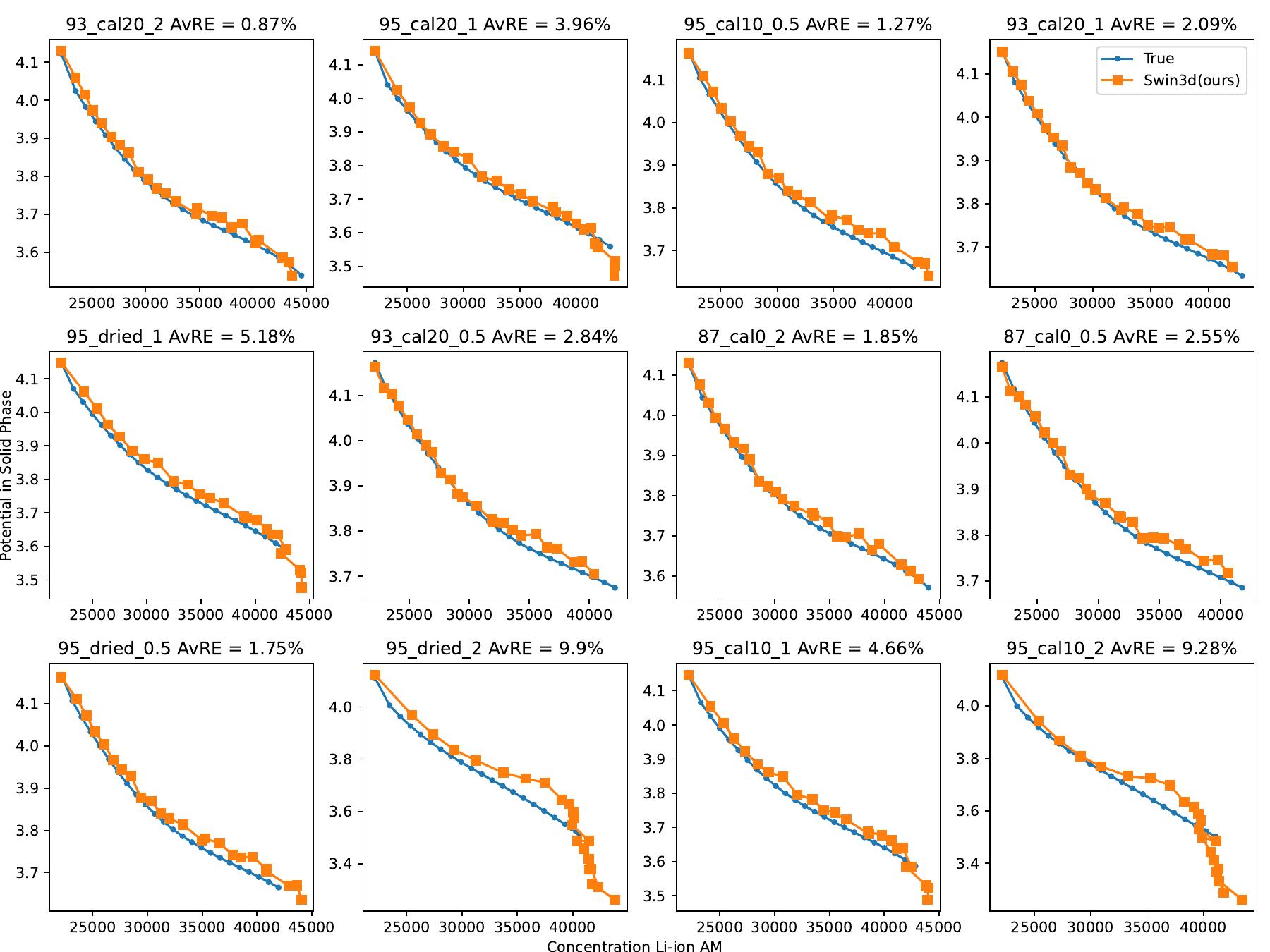}
    \caption{Discharge curves for 12 different sets of electrode parameters.}
    \label{fig:results_curves}
\end{figure}

The visual results demonstrate that the proposed Swin3D architecture significantly outperforms other surrogate models. As shown in the error maps (where red and blue indicate large positive and negative errors, respectively), Swin3D maintains a consistently low error rate across all time steps. Quantitatively, the \texttt{Swin3D\_S\_GPE} variant achieves the lowest Average Relative Error (AvRE) of 0.1974, 0.0914, and 0.1237 at time steps $t = 6, 12, \text{and } 24$, respectively. In contrast, competing models exhibit significantly higher error variance, underscoring the effectiveness of our method in capturing intricate spatial-temporal features.

We further investigate the model's capability to predict discharge behaviors under varying electrode parameters. Fig.~\ref{fig:results_curves} displays the discharge curves for 12 distinct parameter sets.

As shown in Fig.~\ref{fig:results_curves}, the majority of parameter combinations yield an AvRE below 3\%. Outlier cases (\texttt{95\_dried\_0.5}, \texttt{95\_cal10\_2}) exhibit higher errors ($\sim$9--10\%), attributable to a distribution shift where the discharge potential peaks at 3.5\,V rather than the typical 3.6\,V, a limitation to be addressed in future work.

\subsection{Numerical Comparison}

Table~\ref{table:results} presents the quantitative comparison of our method against state-of-the-art point cloud learning baselines, including PointNet \cite{qi2017pointnet}, DGCNN \cite{wang2019dynamic}, Point Transformer \cite{zhao2021point}, and Point Transformer V2 \cite{wu2022point}.

\begin{table}
    \centering
    \caption{Numerical comparison of different surrogate models on the Concentration of Electrochemical Simulation dataset. RMSE: Root Mean Square Error; SMAPE: Symmetric Mean Absolute Percentage Error. Best results are highlighted in bold.}
    \label{table:results}
    \setlength{\tabcolsep}{10pt} 
    \begin{tabular}{lrrr}
        \hline
        \textbf{Methods}   & \textbf{RMSE} $\downarrow$ & \textbf{SMAPE} $\downarrow$ & \textbf{$R^2$} $\uparrow$ \\
        \hline
        PointNet (CVPR17)  & 0.4872                     & 124.5293                    & -0.1633                   \\
        DGCNN (SIGGRAPH19) & 0.2405                     & 92.8054                     & 0.1945                    \\
        PT (ICCV21)        & 0.2348                     & 115.0083                    & 0.3425                    \\
        PTV2 (NeurIPS22)   & 0.4278                     & 120.8603                    & -3.7154                   \\
        \hline
        Swin3D\_L\_GPE     & 0.1535                     & 86.0973                     & 0.5703                    \\
        Swin3D\_M\_GPE     & 0.2140                     & 92.4920                     & 0.4009                    \\
        Swin3D\_S\_GPE     & \textbf{0.0734}            & \textbf{55.4275}            & \textbf{0.7946}           \\
        Swin3D\_T\_GPE     & 0.1269                     & 84.7020                     & 0.6448                    \\
        \hline
    \end{tabular}
\end{table}

The \texttt{Swin3D\_S\_GPE} model achieves the best performance across all metrics, with an RMSE of \textbf{0.0734}, SMAPE of \textbf{55.43}, and $R^2$ of \textbf{0.7946}. Interestingly, the "Small" (S) variant of our architecture outperforms the "Large" (L) and "Medium" (M) variants. This counter-intuitive result can be attributed to the regularization effect of the smaller model capacity. Given the limited size of the electrochemical simulation dataset, the larger models are prone to overfitting, whereas the smaller architecture strikes an optimal balance between feature expressiveness and generalization.

\subsection{Computational Efficiency}

One of the primary motivations for using a deep learning surrogate model is to accelerate the time-consuming physics-based simulations. Table~\ref{tab:runtime} compares the computational costs.

\begin{table}
    \centering
    \caption{Runtime comparison between physics-based Electrochemical Simulation (ES) and AI-driven surrogate models. The ES runtime is for a single simulation instance.}
    \label{tab:runtime}
    \setlength{\tabcolsep}{8pt}
    \begin{tabular}{lcc}
        \hline
        \textbf{Method}          & \textbf{Training Time} & \textbf{Inference Time} \\
                                 & (seconds/epoch)        & (seconds/sample)        \\
        \hline
        Physics-Based Simulation & N/A                    & $\approx$ 46,800 (13 h) \\
        \hline
        PointNet (CVPR17)        & 1.05                   & 0.0130                  \\
        DGCNN (SIGGRAPH19)       & 2.23                   & 0.0279                  \\
        PT (ICCV21)              & 6.36                   & 0.0805                  \\
        PTV2 (NeurIPS22)         & 12.71                  & 0.1569                  \\
        \hline
        Swin3D\_S\_GPE (Ours)    & 11.89                  & 0.1487                  \\
        \hline
    \end{tabular}
\end{table}

The physics-based simulation takes approximately 13 hours (46,800 seconds) to compute a single discharge process. In stark contrast, our \texttt{Swin3D\_S\_GPE} model requires only \textbf{0.1487 seconds} for inference. This represents a speedup of over \textbf{magnitudes of $10^5$}, enabling real-time prediction and high-throughput material screening that were previously computationally infeasible.

\subsection{Ablation Studies}

\subsubsection{Effectiveness of Gaussian Positional Encoding (GPE).}
We evaluate the impact of different positional encoding strategies: standard Cartesian coordinates (XYZ), Sinusoidal encoding (SIN), and our proposed Gaussian Positional Encoding (GPE).

\begin{table}
    \centering
    \caption{Ablation study of positional encoding strategies. Comparisons are made across different model sizes (T, S, M, L).}
    \label{table:ablation_study}
    \setlength{\tabcolsep}{8pt}
    \begin{tabular}{llrrr}
        \hline
        \textbf{Model}     & \textbf{Encoding} & \textbf{RMSE} $\downarrow$ & \textbf{SMAPE} $\downarrow$ & \textbf{$R^2$} $\uparrow$ \\
        \hline
        Swin3D\_S          & XYZ               & 0.0964                     & 69.7779                     & 0.7300                    \\
        Swin3D\_S          & SIN               & 0.0954                     & 72.2204                     & 0.7328                    \\
        \textbf{Swin3D\_S} & \textbf{GPE}      & \textbf{0.0734}            & \textbf{55.4275}            & \textbf{0.7946}           \\
        Swin3D\_T          & GPE               & 0.1269                     & 84.7020                     & 0.6448                    \\
        Swin3D\_M          & GPE               & 0.2140                     & 92.4920                     & 0.4009                    \\
        Swin3D\_L          & GPE               & 0.1535                     & 86.0973                     & 0.5703                    \\
        \hline
    \end{tabular}
\end{table}

\begin{figure}[t]
    \centering
    \includegraphics[width=\textwidth]{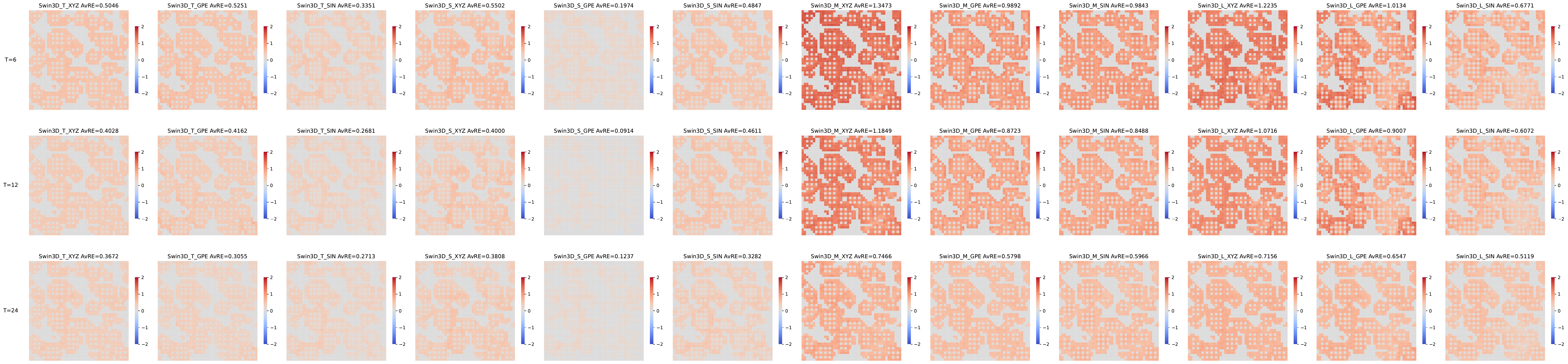}
    \caption{Visual comparison of prediction errors using different positional encodings (XYZ, SIN, GPE). The GPE variant shows reduced error magnitudes in the highlighted regions.}
    \label{fig:loss_xyz}
\end{figure}

As shown in Table~\ref{table:ablation_study}, GPE consistently outperforms XYZ and SIN encodings. For the best-performing `S` model, using GPE reduces the RMSE from 0.0964 (XYZ) to 0.0734. Fig.~\ref{fig:loss_xyz} visually confirms this, showing that models trained with GPE produce smoother and more accurate concentration fields. This suggests that GPE provides a more expressive and adaptive representation for the complex geometries found in electrode microstructures.

\subsubsection{Configuration of Temporal Encoding.}
Finally, we optimized the Temporal Encoding module through a grid search. We explored variations in network depth (1-3 hidden layers), width ($0.5C$ to $2C$), and activation functions (ReLU vs. GeLU). The experimental results indicate that a configuration with \textbf{1 hidden layer}, \textbf{$2C$ neurons}, and \textbf{ReLU activation} yields the best performance. This simple yet effective structure is sufficient to capture the temporal dependencies of the discharge process without introducing unnecessary complexity.
\section{Conclusion}

In this paper, we presented a novel automated framework for integrating deep learning surrogate models into physics-based battery simulation workflows. By leveraging a Swin3D backbone enhanced with Gaussian Positional Encoding and a specialized Temporal Encoding module, our approach effectively captures the complex spatial-temporal dynamics of electrode discharge processes directly from volumetric mesh data.

Empirical results on the Electrochemical Simulation (ES) dataset demonstrate that our model not only outperforms state-of-the-art point cloud baselines in prediction accuracy but also achieves a dramatic reduction in computational time-accelerating simulations from hours to milliseconds. This efficiency breakthrough paves the way for high-throughput material screening and real-time battery management, which were previously computationally prohibitive.

Beyond the specific domain of battery modeling, the proposed pipeline offers a generalized solution for 3D point cloud forecasting tasks in scientific computing. Future work will focus on extending this framework to handle multi-physics coupling scenarios and exploring active learning strategies to further minimize the data dependence of the surrogate models.
\begin{credits}
    \subsubsection{\ackname}
    We thank the reviewers for their insightful feedback and constructive suggestions.

    \subsubsection{\discintname}
    The authors have no competing interests to declare that are
    relevant to the content of this article.
\end{credits}

\bibliographystyle{splncs04}
\bibliography{ref}

\end{document}